\documentclass{article}

\usepackage{PRIMEarxiv}

\usepackage[utf8]{inputenc} 
\usepackage[T1]{fontenc}    
\usepackage{hyperref}       
\usepackage{url}            
\usepackage{booktabs}       
\usepackage{amsfonts}       
\usepackage{nicefrac}       
\usepackage{microtype}      
\usepackage{lipsum}
\usepackage{fancyhdr}       
\usepackage{graphicx}       
\graphicspath{{media/}}     

\title{Understanding the double descent curve in Machine Learning
\thanks{\textit{\underline{Citation}}: 
\textbf{Authors. Title. Pages.... DOI:000000/11111.}} 
}

\author{
  Luis Sa-Couto, Jose Miguel Ramos, Miguel Almeida, Andreas Wichert \\
  Department of Computer Science and Engineering and INESC-ID \\
  Higher Technical Institute, University of Lisbon \\
  Lisbon\\
  \texttt{\{luis.sa.couto, jose.miguel.ramos, miguel.almeida, andreas.wichert\}@tecnico.ulisboa.pt} \\
}

\begin{document}
\maketitle

\begin{abstract}
The theory of bias-variance used to serve as a guide for model selection when applying Machine Learning algorithms. However, modern practice has shown success with over-parameterized models that were expected to overfit but did not. This led to the proposal of the double descent curve of performance by Belkin et al. Although it seems to describe a real, representative phenomenon, the field is lacking a fundamental theoretical understanding of what is happening, what are the consequences for model selection and when is double descent expected to occur. In this paper we develop a principled understanding of the phenomenon, and sketch answers to these important questions. Furthermore, we report real experimental results that are correctly predicted by our proposed hypothesis.
\end{abstract}


\section{Introduction}

\begin{figure}
\centering
\includegraphics[width=0.45\linewidth]{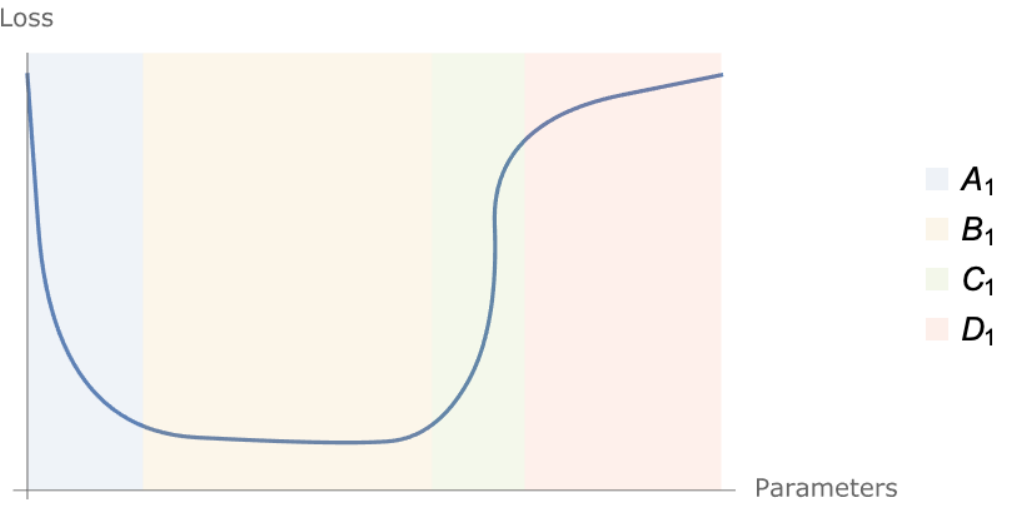}
\includegraphics[width=0.45\linewidth]{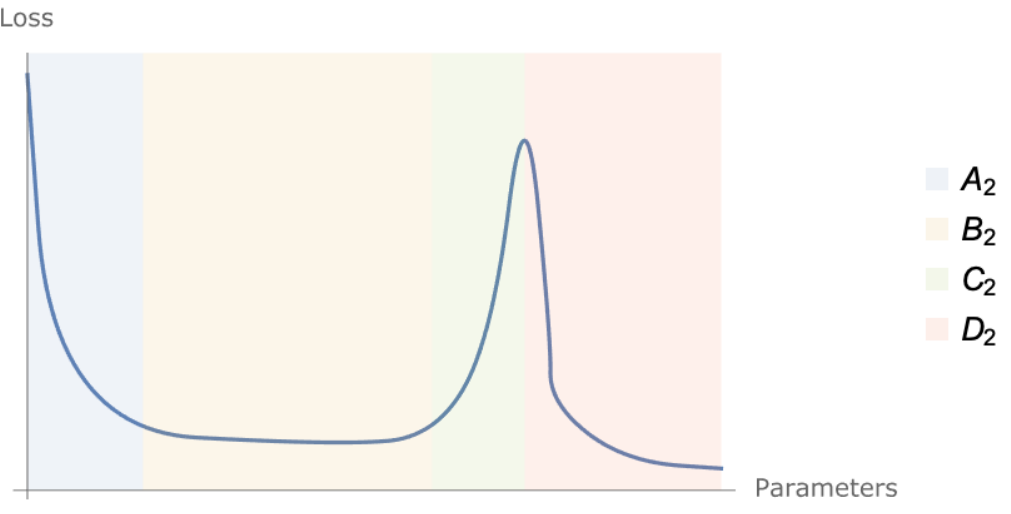}
\caption{On the left we see the scenario that would be predicted by theory, i.e. test set performance keeps degrading as the model becomes more parameterized. On the right we see the double descent phenomenon where over-parameterized models achieve the best generalization.}
\label{hypo}
\end{figure}

The theory of bias variance used to serve as a guide for model selection to try to avoid overfitting \cite{Geman:92,Bishop:06,Prechelt:12,Mei:19}. It used to be thought that one should find the right model complexity, a good balance between under and overfitting \cite{kohavi:96,Yang:20}, i.e. the valley depicted in figure \ref{hypo}.

However, modern practice has shown success with over-parameterized models that were predicted by the theory to overfit but did not \cite{Neyshabur:14,Goodfellow:16,Li:18,Allen-Zhu:19,Allen-Zhu:19a,Du:19,Adavani:20,Oymak:20}. To try to extend theory to meet practice, Belkin et al. \cite{Belkin:19} proposed the double descent curve of performance presented on the right side of Figure \ref{hypo}.

In this curve, two regimes are separated by a theoretical threshold situated at the point where models begin to have more parameters than the number of data points ($N$) multiplied by the number of classes ($K$). While the first regime, under-parameterized, follows bias-variance theory. The second regime, over-parameterized, shows a new descent where test error decreases again possibly to an even lower value than the minimum of the previous regime.

Is this really the case? If so, is the classical theory invalid for the second regime? If not, how can we reconcile it with modern findings about over-parameterized models \cite{Spigler:19}? When does double descent appear and why? In this paper we try to sketch convincing answers to these fundamental questions. A solid understanding of this issue is essential not only to guide future practice, but also to know if the traditional theory needs to be revised \cite{Neyshabur:14,Neal:18,Neal:19}.

\section*{The story of double descent}
The phenomenon of double descent has been shown to occur with many different learning models \cite{Belkin:19,Nakkiran:21}. Yet, in essence, we can break them down into two distinct categories.

On the first one, we place feature-based models where some kind of previously learned or fixed transformation (like random Fourier features, random ReLU codes or random forests) maps the data onto a hidden space, and then a linear output layer is learned on that space. The second group is composed of neural networks, where the hidden transformation and the output layer are learned jointly.

The exact double descent curve, with two regimes separated by the over-parameterization threshold is much easier to observe on the first group \cite{Ba:20} than on neural nets. In spite of that, the particular fact that over-parameterized models achieve top performance is widespread, as evidenced by the success of Deep Learning \cite{Mei:19,Yang:20}.

With this distinction in mind, we will start by investigating feature-based models with a linear output layer. Once we have developed a strong intuition for double descent in those cases, we can extract conclusions for neural networks.

By focusing on feature-based models we can shift all out attention to the learning of the output layer's weights, where the problem is equivalent to a multivariate linear regression.

Following Occam's razor, smooth decision boundaries usually generalize better from train set to unseen data. With this in mind, it is known that, for linear problems, the $l_2$ norm of the weights is a good measure of overfitting. So much so, that techniques like ridge regression minimize it explicitly together with squared error \cite{Krogh:91,Mei:19}.

Intuitively, if we consider one of the output units $f \left( \mathbf{w}^T \mathbf{x} \right) = f \left( w_0 + w_1 x_1 + \cdots w_D x_D  \right)$, a large weight $w_i$ will cause small changes in the input to have a big impact on the final outcome.

Following this intuition, we expect an error vs. performance curve that perfectly follows the theory of bias-variance. At the beginning, the number of features is too small, and so a model is unable to capture the real patterns in the data. In such case, the model is underfitted (regions $A_1$ and $A_2$ in figure \ref{hypo}).

As the number of features grows, the model starts to be able to fit the training data better and better. To do so, weights grow in absolute value. The larger the weight the more impact a small noisy variation of its feature will have on the output. On regions $B_1$ and $B_2$ in figure \ref{hypo}, the model complexity is small enough that it cannot satisfy all the constraints imposed by the training data, so weights are kept under control, and overfitting is bounded.

The closer the number of parameters gets to over-parameterization, the more easily the model satisfies the constraints imposed by the data, and, for that reason, weights increase in absolute value. The more features there are with large weights, the bigger the chance of small noise in any of them to overthrow the output and cause overfitting (regions $C_1$ and $C_2$ in figure \ref{hypo}).

In theory, after the threshold, the problem should only get worse, and we would expect the behavior presented in region $D_1$ of figure \ref{hypo}. However, somehow, this is not at all what seems to happen. Instead, reported results are more in line with region $D_2$ of the same figure. Why is this the case?

\section*{The distributed decision hypothesis}
\label{sec:hypo}
Before diving into the mathematical details behind our proposed hypothesis, let us provide intuition for it.

When the model has a large number of features with large weights, noise in any of them can change the output completely. However, if there is a constraint on the size of the weights, the decision will have to be distributed across features.

To achieve the same influence of a single feature with a large weight, many constrained features will have to ``work together'' by pointing in the same direction. Hence, the output decision will likely be much more robust. Furthermore, with small weights, noise in a restricted amount of features will have almost no impact on the output, thus making the model much less prone to overfitting.

So, is double descent a byproduct of such a constraint? Throughout the rest of this section we will show more formally that this is indeed the case for feature-based models. From that, we can depart to an experimental section that shows our conclusions in practice.

\subsection*{Over-parameterized has a higher chance of finding low norm solutions}
From now on let us call $N$ to the number of data points and $D$ to the number of parameters ($D-1$ features). While learning the linear layer to minimize squared error, one must solve an optimization problem with $D$ variables and $N$ soft constraints of the form:
\begin{equation}
    \left( \mathbf{w}^T \mathbf{x}_i - z_i \right)^2 = 0 \mbox{ for } i=1, \ldots, N,
\end{equation} where $\mathbf{x}_i$ is the $i$-th data point, $z_i$ is its target, and $\mathbf{w}$ is the weight vector.

\begin{figure}
\centering
\includegraphics[width=0.45\linewidth]{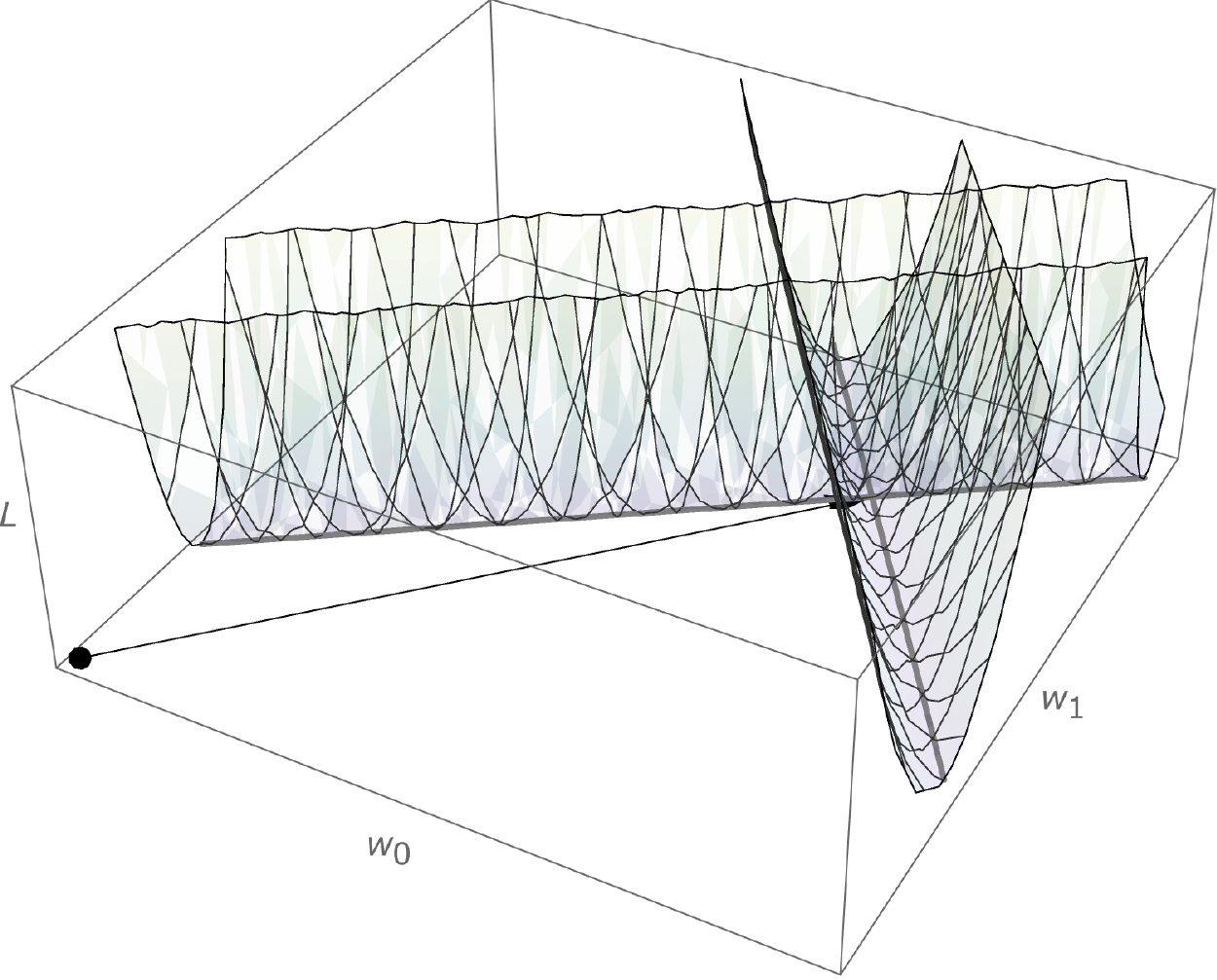}
\includegraphics[width=0.45\linewidth]{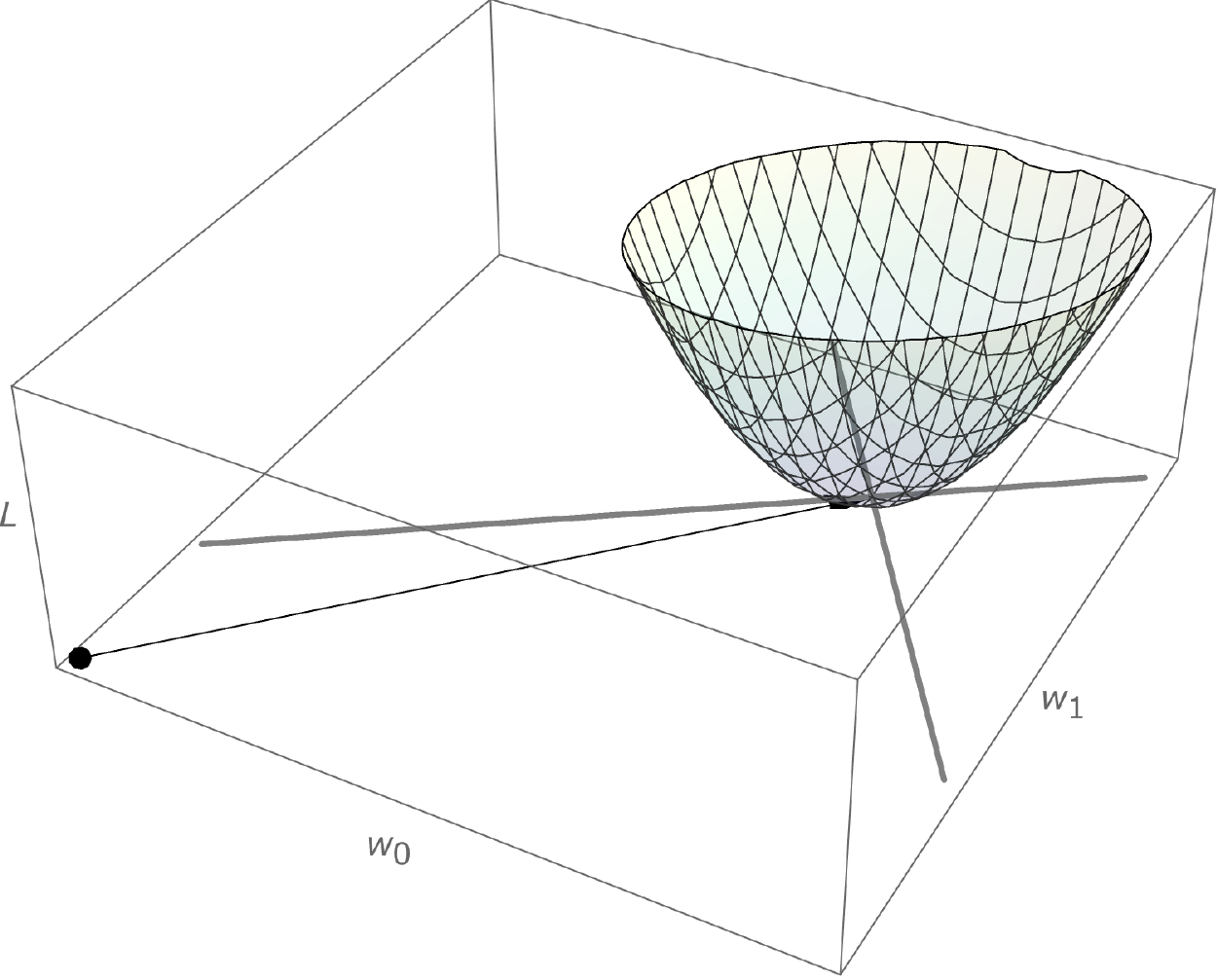}
\caption{On the left we see the 3D energy surfaces induced by a pair of data points on the weight space. On the right we see the sum that yields the final loss function. In this scenario, a single point minimizes the squared error.}
\label{overdetermined3D}
\end{figure}

Assuming the data is linearly independent, while $N \leq D$ the problem is over-determined and a single solution can exist. To provide a visual intuition for it, figure \ref{overdetermined3D} presents the $N=D=2$ case where we can see that the two constraints yield a convex bowl with a single minimum. So, good or bad, the outcome of learning will have a specific $l_2$ norm, and, consequently will incur in a specific amount of overfitting.

\begin{figure}
\centering
\includegraphics[width=0.45\linewidth]{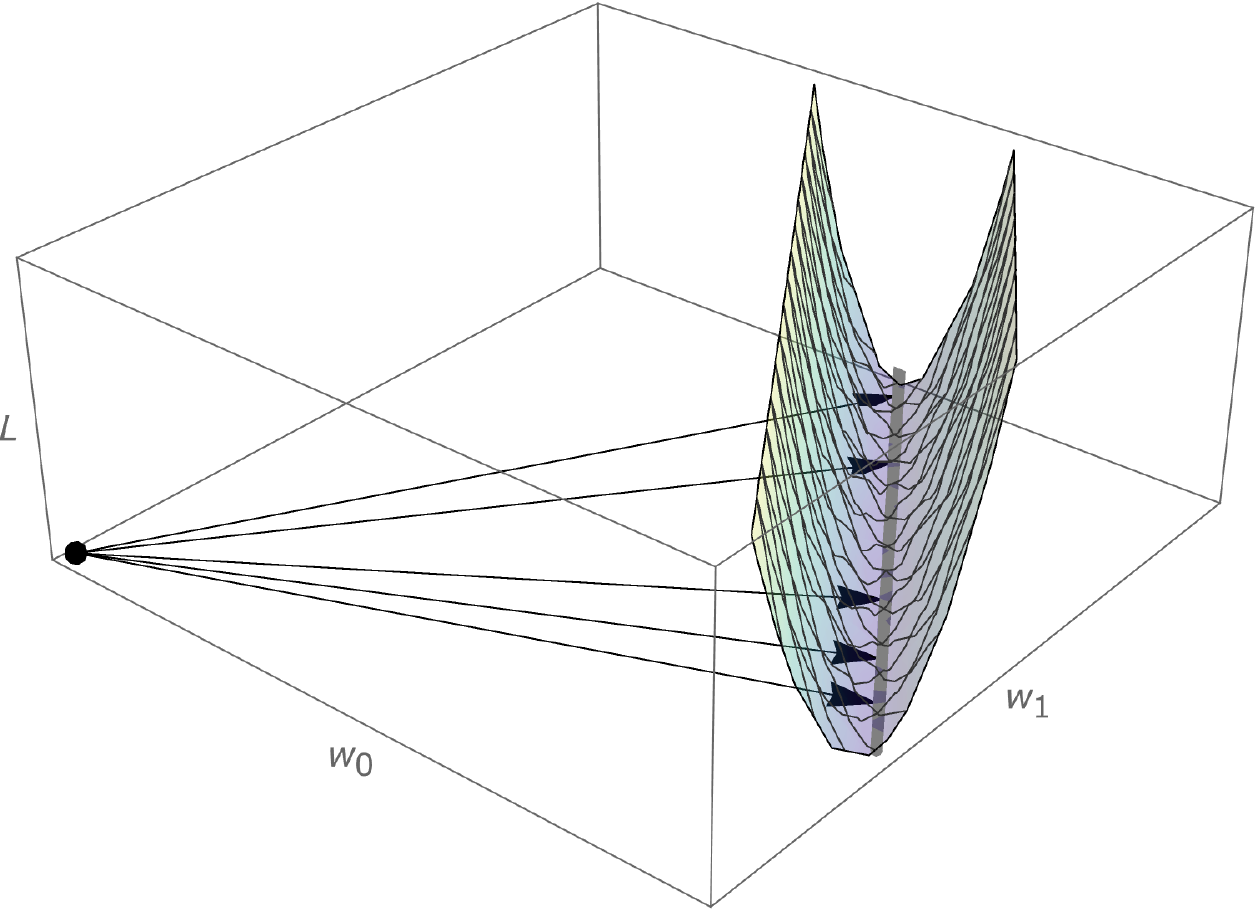}
\caption{With a single data point constraint on a problem with two weights, the system will be over-determined, and the infinite solutions lie on an $D-N$ hyperplane, which in this case is a line.}
\label{underdetermined3D}
\end{figure}

As we reach the over-parameterized regime (when $D>N$) the optimization problem can be equally solved by an infinite number of solutions. More concretely, a $D-N$ dimensional hyperplane of possible weight vectors. To get a visual sense for it, figure \ref{underdetermined3D} shows the $D=2,N=1$ case. There, we see a single valley constraint with an infinite number of equally valid solutions on a line.


Assuming the size of the data set $N$ is fixed, for an increasing number of parameters $D$, the passage from under- to over-parameterized will get us from a singleton solution set $U = \{ \mathbf{w} \}$, to an infinite set of solutions $O = \{ \mathbf{w}_1,\mathbf{w}_2,\mathbf{w}_3,\ldots \}$ to the learning problem. 

\begin{figure}
\centering
\includegraphics[width=0.4\textwidth]{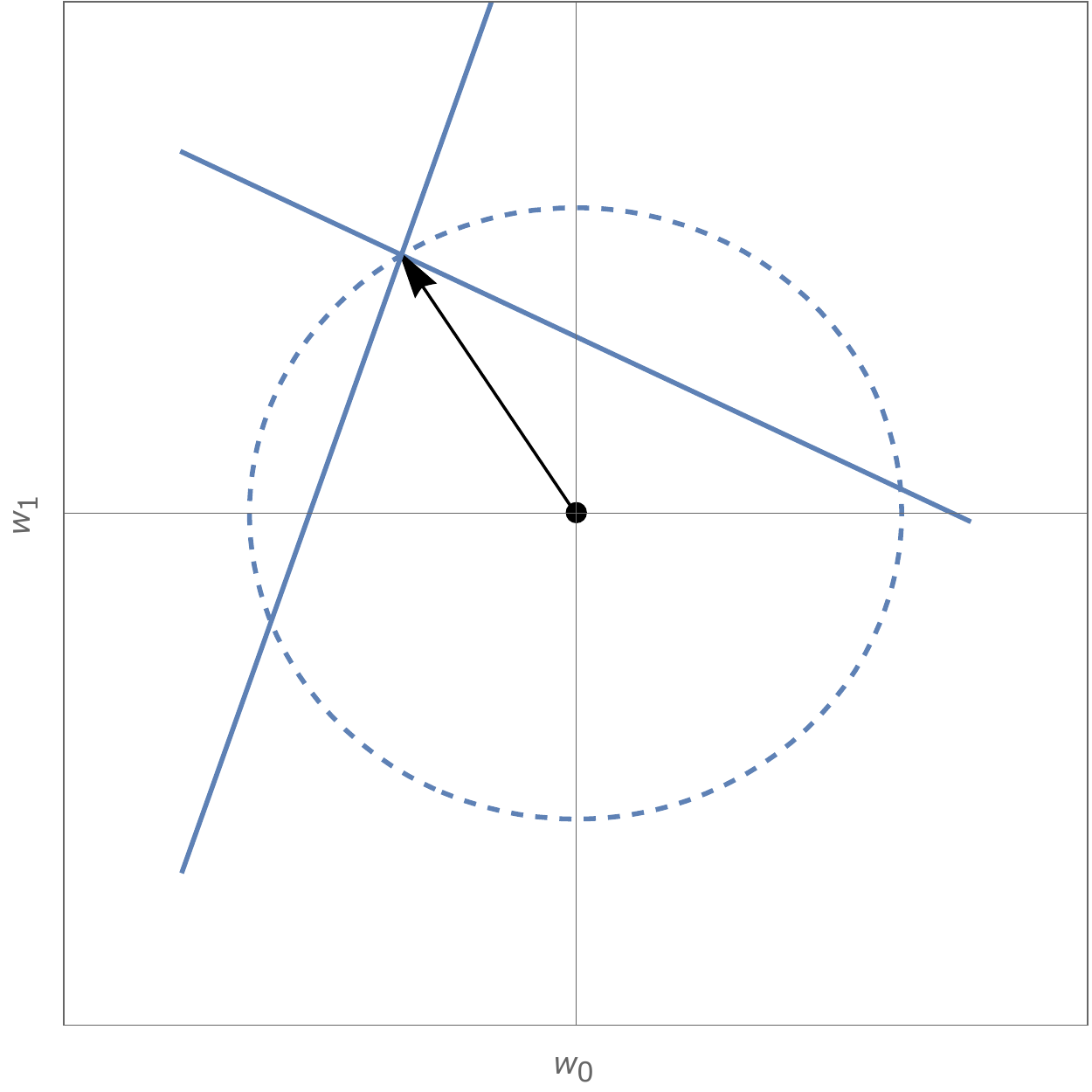}
\includegraphics[width=0.4\textwidth]{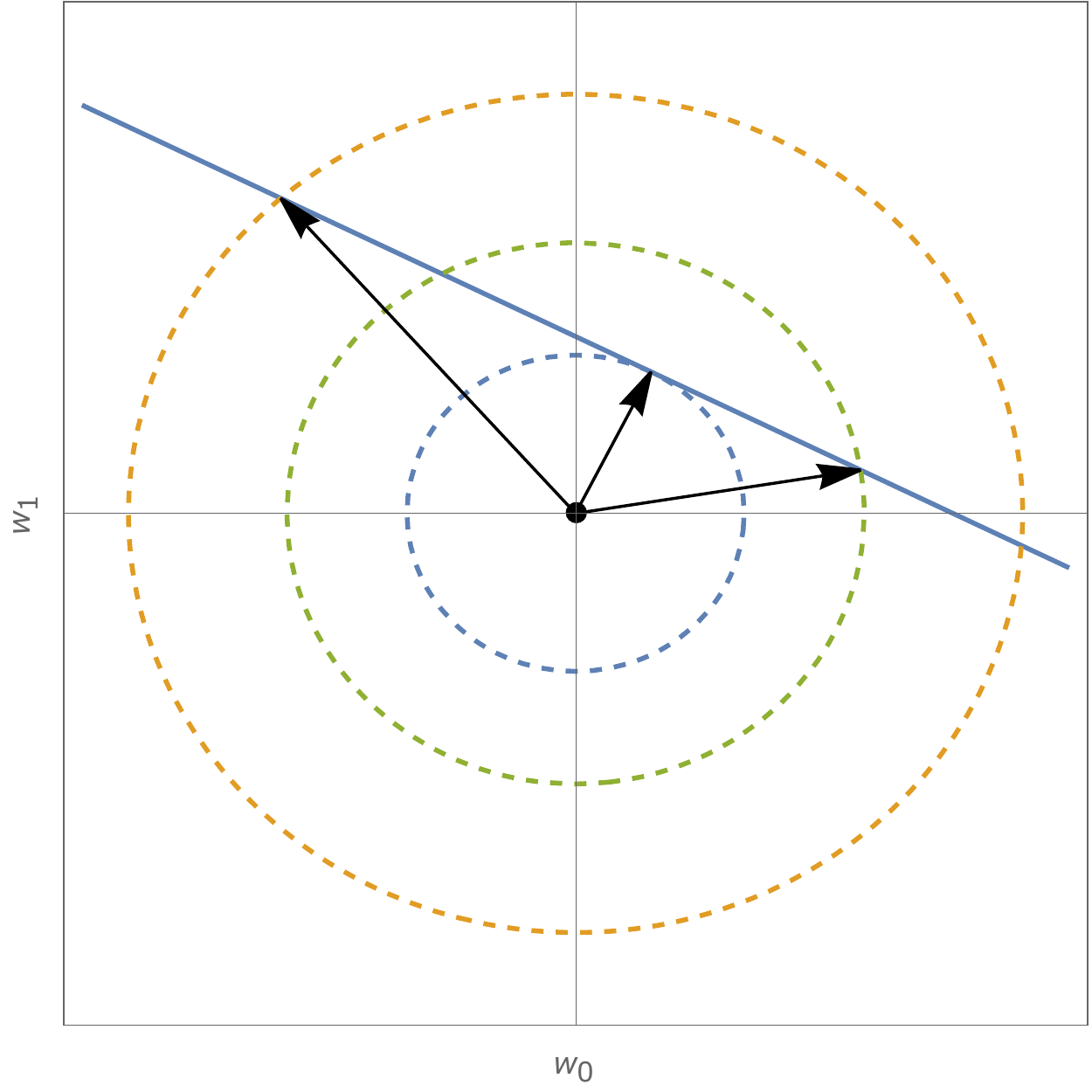}
\caption{Projecting both the under- and the over-determined cases to two dimensions, we see that, to minimize training error, the former is forced to a solution with a fixed $l_2$ norm, whereas the latter can be solved by infinite solutions, with different lengths. Hence, the probability of finding a low norm solution is higher on the over-parameterized scenario.}
\label{1solutionVsMany}
\end{figure}

Now, it is extremely likely that there exists some solution in the infinite set that has a smaller norm than the single solution of the under-parameterized case. More concretely, it is very probable that $\exists \mathbf{w}_o \in O : \| \mathbf{w}_o \|_2 \leq \| \mathbf{w} \|_2$. This idea is visually depicted in figure \ref{1solutionVsMany}, where we project figures \ref{overdetermined3D} and \ref{underdetermined3D} to two-dimensions using contour plots.

When the solution is not unique, the learning mechanism must somehow untie between equally valid minima. Hence, if learning contains any bias towards small norms, the over-parameterized case will, with a very high probability, output the minima that possesses the smallest norm. Therefore, this minima will likely generalize much better to unseen data, as we have previously stated that the weight norm is a good predictor of overfitting.

So, is there such bias in the experiments that have shown double descent? In the experiments section we will show that the answer to this question is a clear yes. In fact, for the over-parameterized case, ridge regression is typically used, where a penalty that depends on the $l_2$ norm of the weight is added to the loss function.

At this point we have established, a qualitative difference between the under-parameterized case ($D \leq N$) and the over-parameterized case ($D > N$), and it explains why generalization error reduces after the threshold (i.e. the second descent). Yet, it does not entirely explain why modern practice has found that the more over-parameterized the model, the better the results seem to get. To explain this is the goal of the next subsection.

\subsection*{The more over-parameterized the better}

\begin{figure}
\centering
\includegraphics[width=0.9\textwidth]{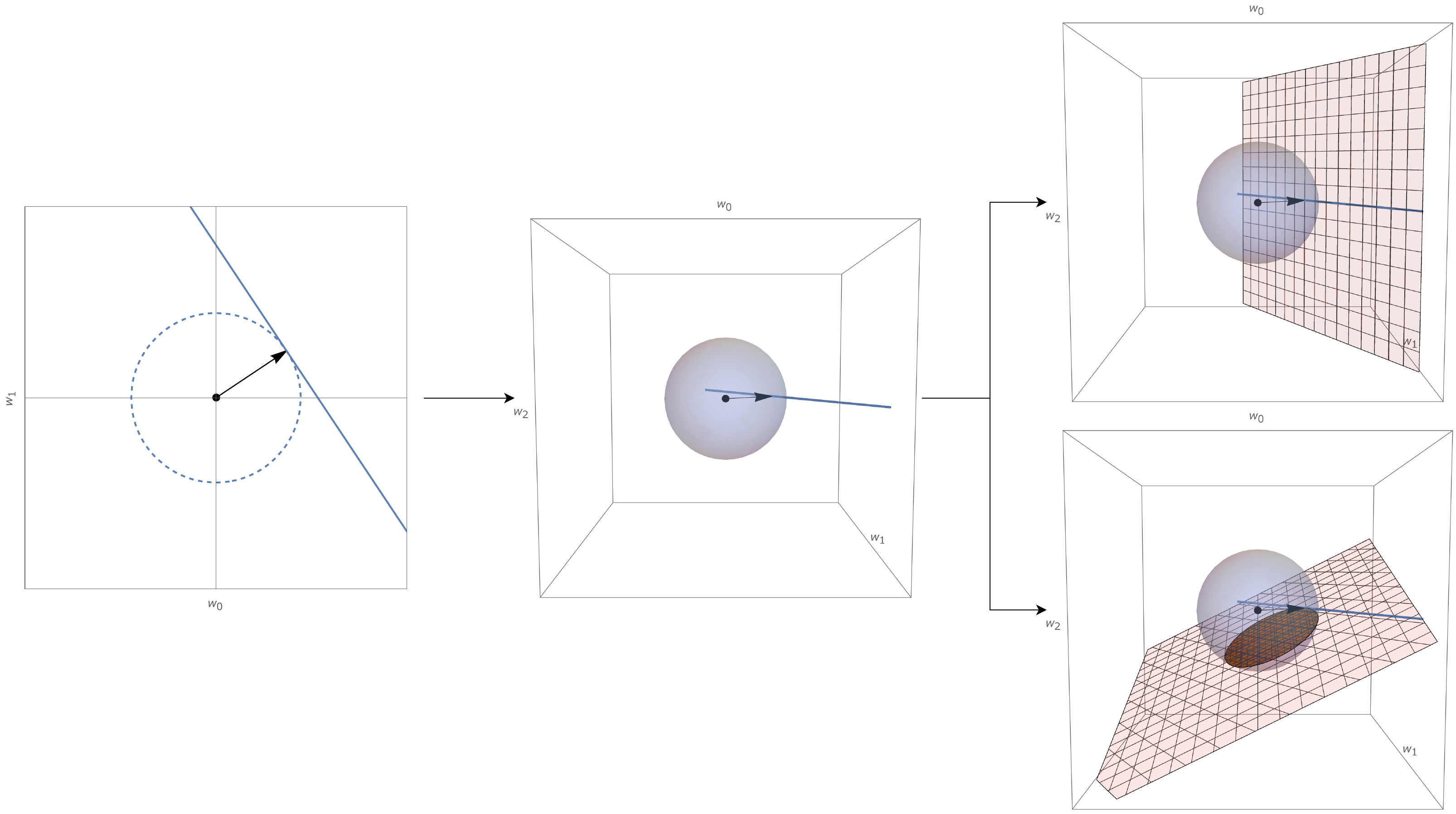}
\caption{For a fixed number of data point constraints (in this case $N=1$), as the number of weights grows, the new hyperplane of solutions will either be tangent or secant to the smallest $l_2$ ball that contains the solution. On the former case, the new solution will have the same norm, whereas on the latter case the solution will have a lower norm. Therefore, as the number of weights grows, the solution norm decreases non-monotonically.}
\label{growD}
\end{figure}

Let us take the over-parameterized case we depicted in figures \ref{underdetermined3D} and \ref{1solutionVsMany}, and see what happens as we add another parameter. Unfortunately, we can no longer plot the error landscape as it is four-dimensional. Yet, we can still plot the weight space and the region of solutions. We do so in figure \ref{growD}.

The previous solution has a given $l_2$ norm. This norm forms a ball $B$ around the origin that contains all the weight vectors that have an equal or smaller $l_2$ norm. Ideally, a better solution, would achieve the same training error while being inside the ball. If that was the case, the new solution would have a smaller norm, and, thus, presumably better generalization capabilities.

With the new parameter, the region of minimas will be a plane that contains the line of solutions to the previous problem. Now, since it contains the previous line of solutions, this plane is guaranteed to the touch the ball $B$. 

Furthermore, a plane that touches a ball can either be tangent to it, or secant. On the former case, the new solution will be the same as the previous one, and the new weight will equal zero. However, much more likely is the latter case where the plane is secant. In this scenario, all the highlighted cross-section of $B$ will contain solutions that are presumably better than the previous one.

This reasoning can be extended to more dimensions, and thus we can conclude that, in the over-parameterized regime, if learning is biased for small norms, as the dimensionality increases, the solution norm is guaranteed to decrease non-monotonically.

\subsection*{Consequences for neural networks}
In neural networks, there is not a smooth transition between one global minima and an infinity of them at the over-parameterization threshold. In fact, these networks can have lots of local minima even when the model is under-parameterized. For that reason, if our view is correct, we would not expect the theoretical threshold of $N \times K$ to predict the regime change. We will test this prediction in the experimental section.

Another prediction that follows relates to the small norms constraint. According to our view, in experiments where double descent happens the learning probably contains some bias to small norms. This bias can be explicit, in the form of weight decay \cite{Krogh:91}. Or it can be implicit, for instance, if we start the weights with small values, and then run a stochastic gradient descent with some form of early stopping, we will be biasing the model to weights in the neighborhood around the origin, and, therefore, to low norm solutions.

\section*{Experiments}

\subsection*{Results on feature-based models}
The double descent phenomenon was shown \cite{Belkin:19} in multiple different feature based models on different data sets. Yet, all share the same pattern. A first feature extraction step transforms the data set, and then a linear layer (i.e. a multivariate linear regression) is applied to solve the supervised task at hand.

To implement the first step, the original work, tried Random Fourier features, Random Forests, and Random ReLU features. In this section for simplicity purposes, and without loss of generality, we will focus our experiments on the latter. Furthermore, we will use a subset of the MNIST data set of handwritten digits \cite{Lecun:98} composed by $3000$ images of $28\times28=784$ pixels, with $10$ possible classes.

\subsubsection*{Over-parameterized is only better with a bias for small norms}
With a fixed feature extraction step, learning amounts to solving the linear regression equations for the output layer.
So, assuming $N$ training examples and $K$-dimensional outputs, the over-parameterization threshold sits at $NK$. Before this point, the system is under-parameterized, and a single global minimum exists. Afterwards, infinite minima appear.

According to our hypothesis, the only reason why performance increases after the threshold is due to bias in learning that prefers weight vectors with small norms. Therefore, we can clearly state a prediction, and design an experiment to test it.

\paragraph{Prediction:} in the over-parameterized regime, if we choose small norm solutions, double descent will appear. Yet, if we choose equally valid higher norm solutions it will not.

\paragraph{Experiment:} when learning the weights, minimize the squared error, but add a penalty that pulls the solutions towards a $D+1$ dimensional point $\mathbf{p}$ of norm $R = \| \mathbf{p} \|_2$. Specifically, fix a small $\lambda$ (i.e. $10^{-8}$) and optimize
\begin{equation}
    L\left( \mathbf{w} \right) = \sum_{i=1}^{N} \left( \mathbf{w}^T \mathbf{x}_i - z_i \right)^2 + \lambda \sum_{d=0}^{D} \left( \mathbf{w}_d - \mathbf{p}_d \right)^2.
\end{equation}
Solve this problem multiple times, for different values of $R$, and collect the training and test squared errors for each of them as the number of parameters increases.

\begin{figure}
\centering
\includegraphics[width=0.45\linewidth]{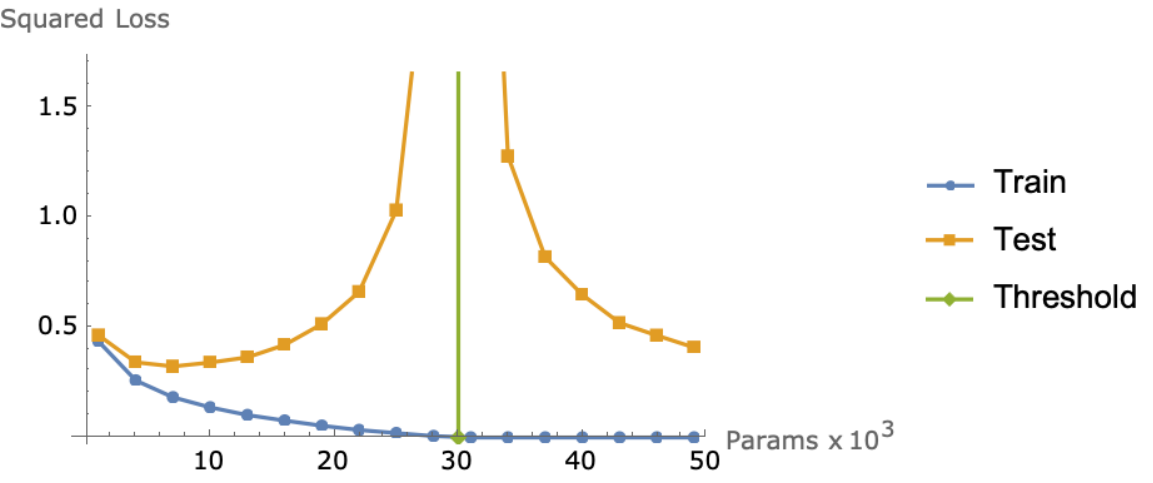}
\caption{Exact reproduction of the double descent curve on a feature-based model.}
\label{LR_DoubleDescent}
\end{figure}

\begin{figure}
\centering
\includegraphics[width=0.45\textwidth]{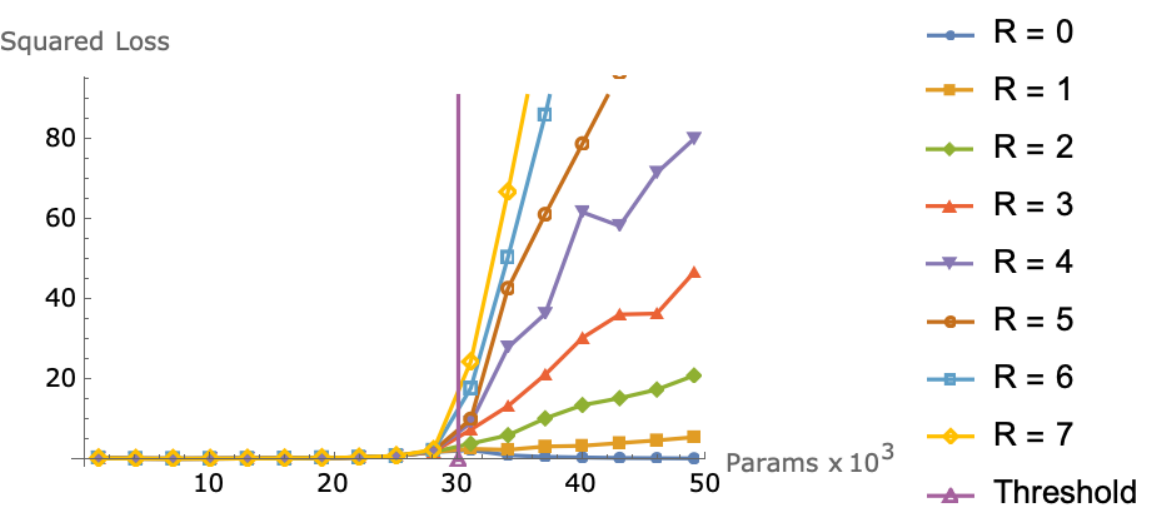}
\includegraphics[width=0.45\textwidth]{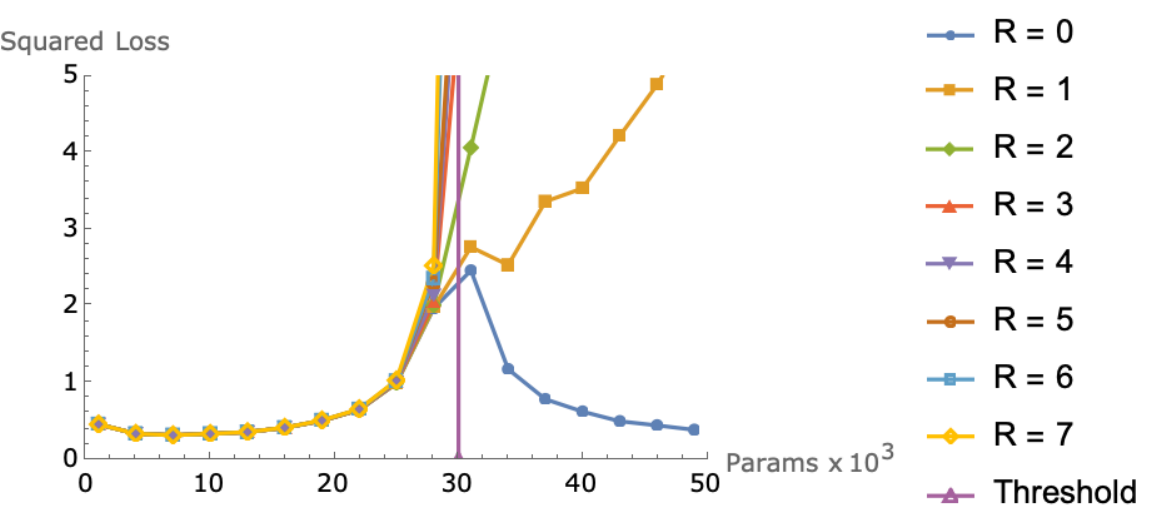}
\caption{Instead of biasing the over-parameterized learning towards the origin, we bias it to some random point $p$ that sits at an Euclidean distance of $R$ from the origin. On the left we have a global view of the plot, whereas on the right we provide a zoom in for smaller loss values. When $R=0$ we get double descent. However, as we increase $R$ performance deteriorates and we get closer and closer to what is predicted by bias-variance theory.}
\label{LR_ChangeRegularizer}
\end{figure}

Figure \ref{LR_DoubleDescent} shows the results for a low norm constraint, whereas figure \ref{LR_ChangeRegularizer} presents them for higher norm constraints. Observation exactly meets expectations. On the one hand, with restrained weights, double descent appears, and over-parameterized test performance is good. On the other hand, choosing equally deep minima with higher norms, the post threshold regime is even worse, just like bias-variance would predict.

\subsubsection*{Enough constraint makes the difference between the two regimes vanish}
If $\lambda$ increases, eventually, the solution weights will not be a minimum of squared error, and only low norm solutions will be returned. With that in mind, another prediction follows.

\paragraph{Prediction: }  for large enough $\lambda$ the difference between under- and over-parameterized regimes should vanish.

\paragraph{Experiment: } solve the multivariate regression equations for increasing values of $l_2$ regularization. For each $\lambda$, collect the training and test losses as the number of features grows.

\begin{figure}
\centering
\includegraphics[width=0.45\textwidth]{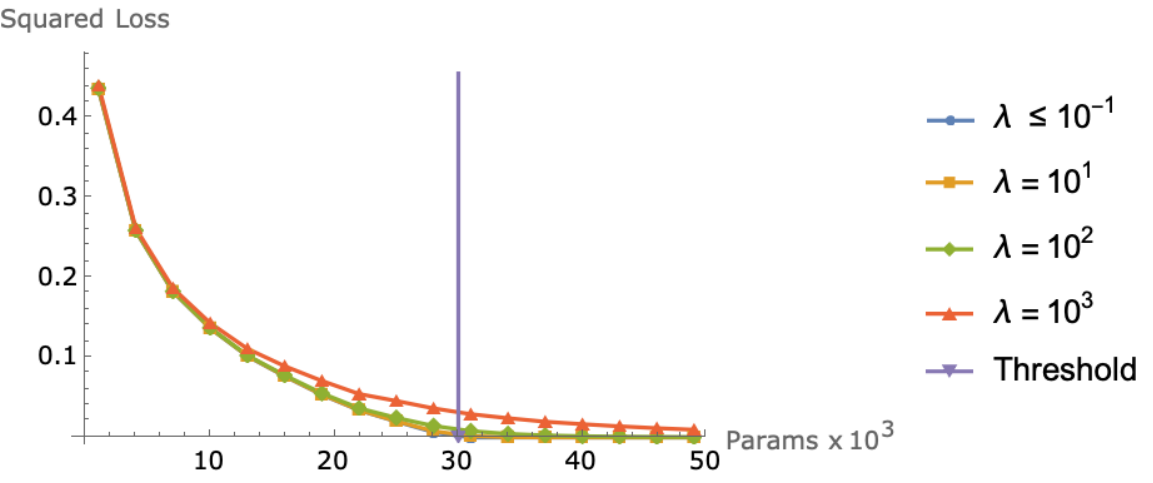}
\includegraphics[width=0.45\textwidth]{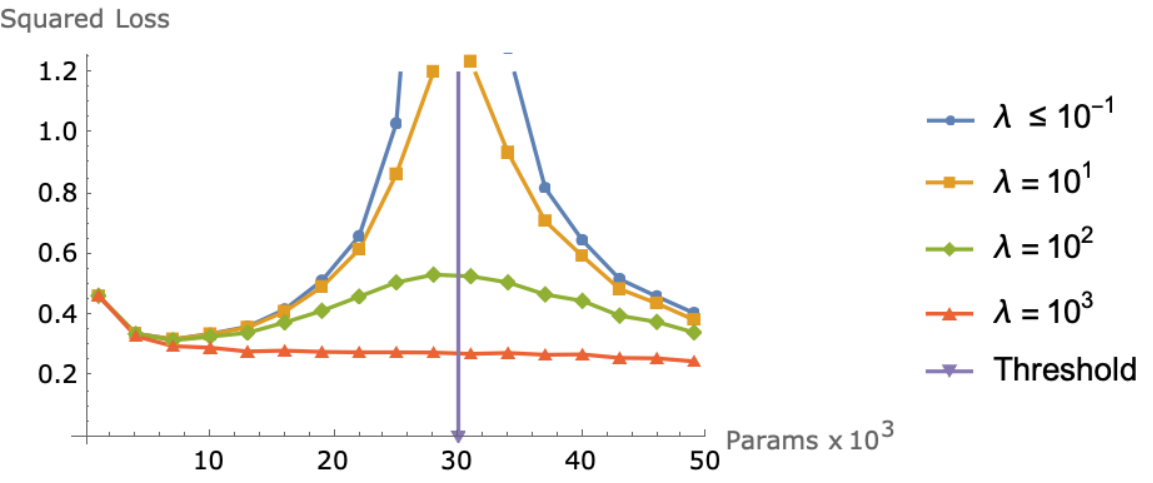}
\caption{On the left we present the training losses for different regularization strengths, whereas on the right we present the test losses. As the bias for small norms gets stronger, the learning process starts to place less value on the data point constraints. Although training error increases, the resulting weight vectors have small norms, and thus generalization is much better throughout. The difference between under- and over-parameterized fades away.}
\label{LR_ChangeLambda}
\end{figure}

In figure \ref{LR_ChangeLambda} we present the error vs. complexity curves for increasing values of the weight penalty $\lambda$. For small $\lambda$, double descent appears. As the penalty increases we see the overfitting peak disappear, and double descent becomes a single descent.

In conclusion, our explanation makes correct predictions about experimental data. In fact, the hypothesis is so well grounded that we expect it to fail only in a very special case where high $l_2$ norm of the weight vector does not predict overfitting. Next we turn to the less predictable case of neural networks.

\subsection*{Results on neural networks}
In Belkin et al \cite{Belkin:19}, to be able to show the phenomenon of double descent in neural networks, the authors use an experimental methodology that involves weight reuse as the complexity of the network being trained increases.

More specifically, they start by creating a network with a hidden layer of size $h$ and training it over a large number of epochs. Then, collect the losses for train and test set. Afterwards, when repeating the experiment for a larger network (i.e. hidden layer of size $h+1$), the authors do not create it from scratch. Instead, they take the previous network, with already trained weights, and add a randomly initialized neuron to the hidden layer. By doing so, the new network will start training at a very advanced stage.

This weight reuse strategy is employed up until the proposed over-parameterization threshold. After which, networks are created from scratch and, thus, initialized with a random-like strategy.

\subsubsection*{The second descent is a product of the early stopping constraint}
When performing experiments with Neural Networks, the decision on when to stop training has a big impact on performance. Early stopping works as a kind of regularizer that avoids an explosion in the network's weights \cite{Neyshabur:14} that can happen with late stopping.

In the weight reuse methodology, networks before the threshold start their learning at a later stage, so it is as if they are being trained in a late stopping regime. After the threshold, networks are being comparatively stopped early. This difference is enough to explain the sudden drop in test error that is reported.

\paragraph{Prediction: } if we do not early stop the networks after the threshold, but instead keep reusing the weights, then the over-parameterized models will be just as bad or even worse.

\paragraph{Experiment: } replicate the aforementioned methodology, without turning off weight reuse after the threshold.

\begin{figure}
\centering
\includegraphics[width=0.45\linewidth]{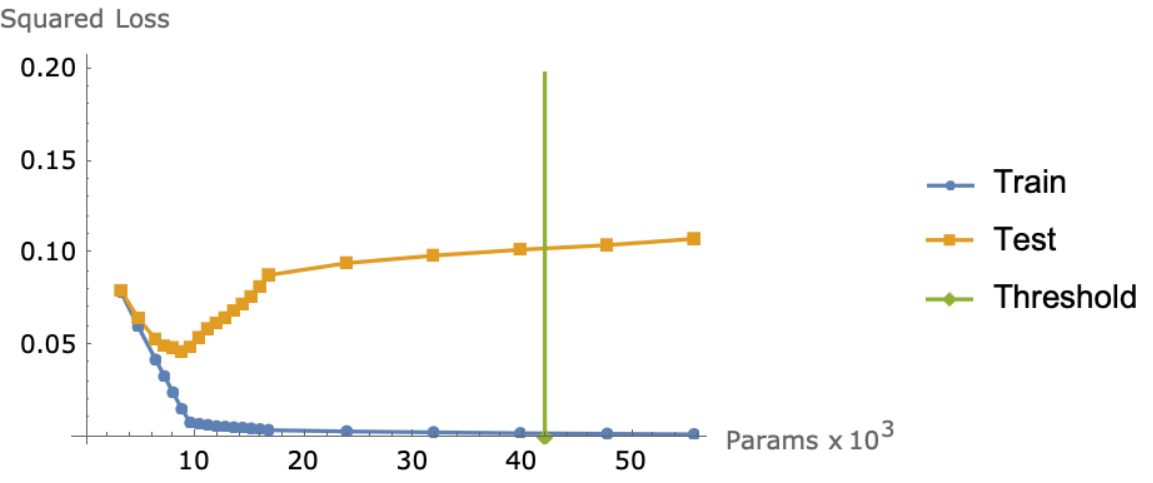}
\caption{By not shutting down the weight reuse scheme at the threshold, all networks are late stopped, so the weight constraint is diminished, and we get bad over-parameterized performance.}
\label{NN_NoDescent}
\end{figure}

We applied the proposed experiment on a MNIST subset with $4200$ images. The results are reported in Figure \ref{NN_NoDescent}. Analyzing them, we see a clear alignment with our intuition. Namely that without the early stopping constraint, performance degrades even further. Additionally, passing the threshold seems to have no impact.

\subsubsection*{The threshold does not predict the regime change}
Our analysis of feature-based models led to the conclusion that the over-parameterization threshold is key because it takes the learning problem from one with a single global minima, to another with an infinite number of them. This idea does not apply to neural networks that can have multiple local minima from the start. Putting this together with the previous results we can generate another prediction.

\paragraph{Prediction: } the over-parameterization threshold does not predict the regime,  the change in the initialization scheme causes an illusory regime change.

\paragraph{Experiment: } take the same MNIST subset and replicate the original paper's experiment, but vary the point at which the reuse scheme is switched off. 

\begin{figure}
\centering
\includegraphics[width=0.45\textwidth]{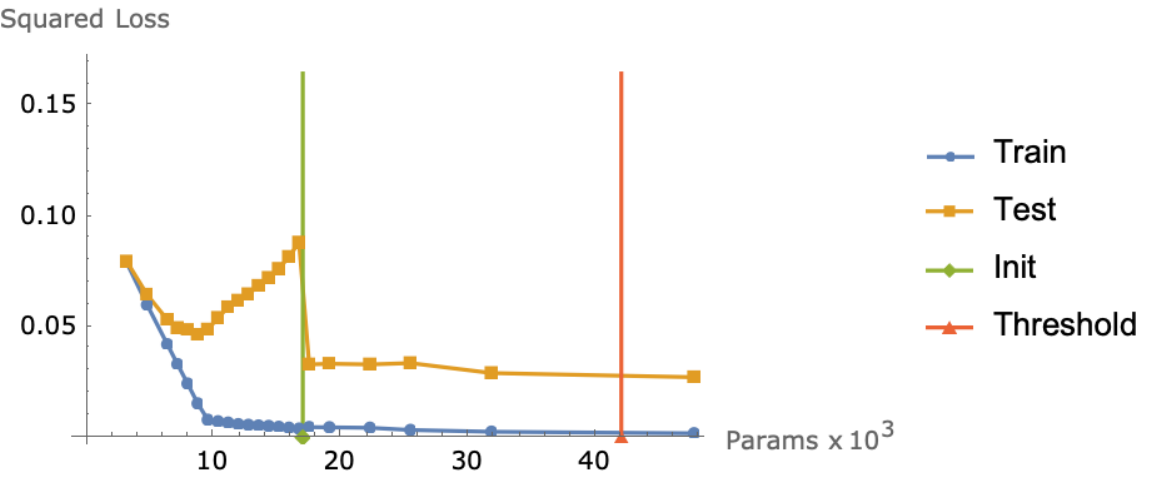}
\includegraphics[width=0.45\textwidth]{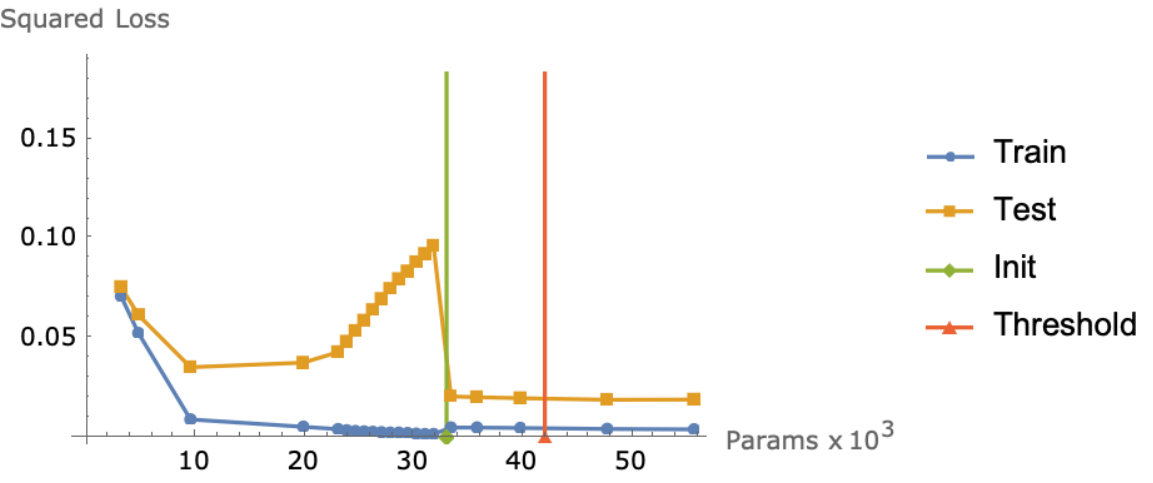}
\includegraphics[width=0.45\textwidth]{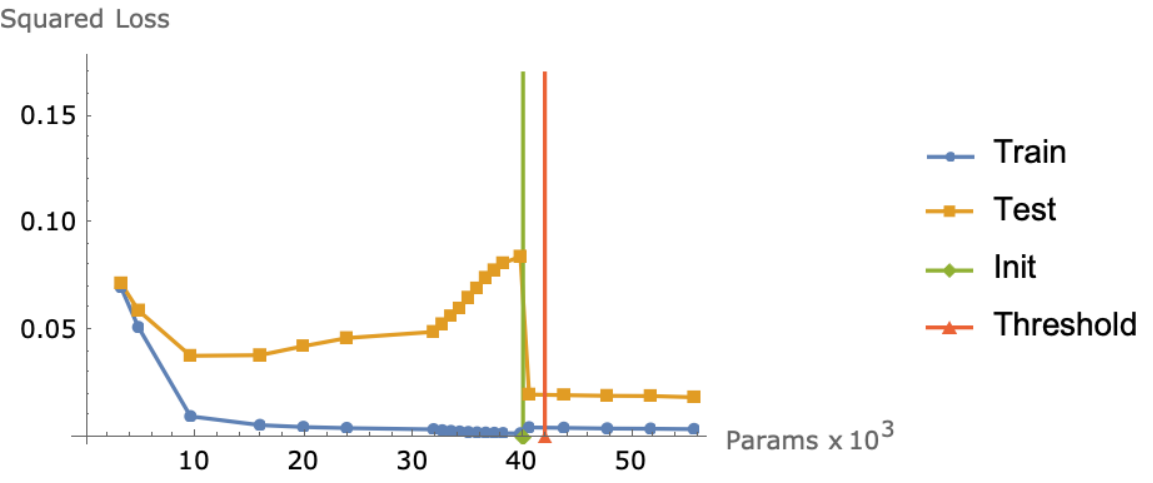}
\caption{The regime change happens exactly at the point where the initialization scheme changes, and not at the theoretical threshold.}
\label{NN_Threshold}
\end{figure}

In Figure \ref{NN_Threshold} we present three performance curves where the switch happened at different model complexity levels. We can immediately see that the so called second descent, occurs at this point, and not at the threshold.

\subsubsection*{Over-parameterized with constraint is best}
Modern practice points to the fact that best results are attained, when large networks are used. Generalizing the intuitions from section \ref{sec:hypo}, we would expect that these successful over-parameterized networks are being constrained somehow, thus creating distributed decisions.

\paragraph{Prediction: } if we take over-parameterized networks and, somehow, constrain their weights, the better results will come from the more parameterized ones.

\paragraph{Experiment: } initialize a network with small random weights, and train it with early stopping. Collect train and test losses, then repeat the process for larger and larger networks.

\begin{figure}
\centering
\includegraphics[width=0.445\linewidth]{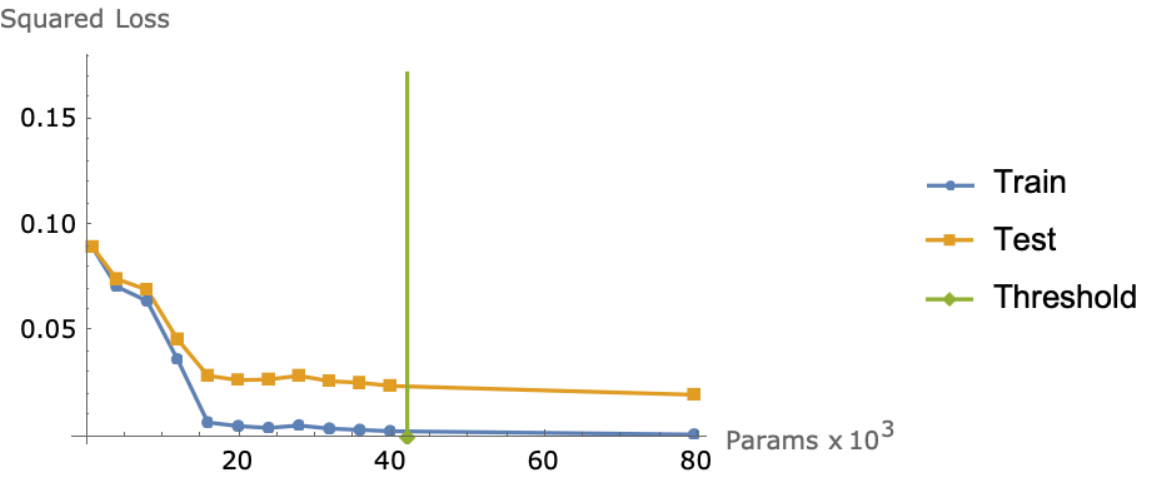}
\caption{Increasingly complex networks are trained with early stopping. As complexity increases, performance tends to improve.}
\label{NN_Biased}
\end{figure}

The experimental results are reported in figure \ref{NN_Biased}. The resulting error curve not only matches what is reported throughout modern practice, but also confirms the intuition that over-parameterized with weight constraint is best. In fact, it is reminiscent of the previously presented experiment with highly regularized feature-based models. 

\section*{Conclusion}
Starting from the double descent story, we decided to investigate the setting where it most clearly appears: feature-based models. In this case, learning amounts to solving a multivariate linear regression on a fixed hidden space.

In this learning problem, the over-parameterization threshold $N \times K$ plays a key role. Before it, the optimization problem has a single minimum. After it, the problem has infinite, equally valid global minima that sit on a hyperplane of dimension at most $D-N$.

With many solutions, some bias needs to break the tie. Typical experiments choose the lowest norm weight vector, as it is a good predictor of overfitting, and this makes all the difference. This constraint is responsible for the good behavior on the post-threshold regime, and, without it, the regime follows the classical predictions of bias-variance. Furthermore, we have shown that, with the constraint, the higher the number of parameters the more likely it is for the solution to improve, which helps explain modern observations on the success of large models.

Afterwards, using our findings on feature-based models, we showed that the idea of a regime change at an over-parameterization threshold cannot be directly exported to neural networks because the number of minima does not go from one to infinity. However, assuming that the number of minima grows with the increase of complexity, we intuit that, with some overfitting constraint large models will be better.

With that in mind, we explained how initializing weights on a small norm region, and running stochastic gradient descent with early stopping can play the role of the norm constraint. With this setting, we explained why, unlike what bias-variance theory would predict, over-paramaterized neural networks perform well.

In conclusion, although bias-variance theory is valid, and modern practice seems to contradict it, the latter is entirely justified and well-grounded. Specifically, provided that we can impose some bias towards less overfitted solutions, we should use over-parameterized models.

\section*{Acknowledgments}
We would like to acknowledge support for this project
from the Portuguese Foundation for Science and Technology (FCT) with a doctoral grant SFRH/BD/144560/2019 awarded to the first author, and the general grant UIDB/50021/2020. The Foundation had no role in study design, data collection and analysis, decision to publish, or preparation of the manuscript. The authors declare no conflicts of interest. Code and data for all the experiments can be obtained by email request to the first author.

\bibliographystyle{unsrt}  
\bibliography{paper}

\begin{thebibliography}{10}

\bibitem{Geman:92}
Stuart Geman, Elie Bienenstock, and René Doursat.
\newblock {Neural Networks and the Bias/Variance Dilemma}.
\newblock {\em Neural Computation}, 4(1):1--58, 01 1992.

\bibitem{Bishop:06}
C.~M. Bishop and N.~M. Nasrabadi.
\newblock {\em Pattern recognition and machine learning.}, volume~4.
\newblock Springer, 2006.

\bibitem{Prechelt:12}
Lutz Prechelt.
\newblock {\em Early Stopping --- But When?}, pages 53--67.
\newblock Springer Berlin Heidelberg, Berlin, Heidelberg, 2012.

\bibitem{Mei:19}
Song Mei and Andrea Montanari.
\newblock The generalization error of random features regression: Precise
  asymptotics and double descent curve.
\newblock {\em arXiv}, 2019.

\bibitem{kohavi:96}
Ron Kohavi, David~H Wolpert, et~al.
\newblock Bias plus variance decomposition for zero-one loss functions.
\newblock In {\em ICML}, volume~96, pages 275--83, 1996.

\bibitem{Yang:20}
Zitong Yang, Yaodong Yu, Chong You, Jacob Steinhardt, and Yi~Ma.
\newblock Rethinking bias-variance trade-off for generalization of neural
  networks.
\newblock {\em arXiv}, 2020.

\bibitem{Neyshabur:14}
T.~Ryota B.~Neyshabur and N.~Srebro.
\newblock In search of the real inductive bias: On the role of implicit
  regularization in deep learning.
\newblock {\em arXiv}, 1412(6614), 2014.

\bibitem{Goodfellow:16}
Ian Goodfellow, Yoshua Bengio, and Aaron Courville.
\newblock {\em Deep Learning}.
\newblock MIT Press, 2016.

\bibitem{Li:18}
Yuanzhi Li and Yingyu Liang.
\newblock Learning overparameterized neural networks via stochastic gradient
  descent on structured data.
\newblock {\em Advances in neural information processing systems}, 31, 2018.

\bibitem{Allen-Zhu:19}
Zeyuan Allen-Zhu, Yuanzhi Li, and Yingyu Liang.
\newblock Learning and generalization in overparameterized neural networks,
  going beyond two layers.
\newblock {\em Advances in neural information processing systems}, 32, 2019.

\bibitem{Allen-Zhu:19a}
Zeyuan Allen-Zhu, Yuanzhi Li, and Zhao Song.
\newblock A convergence theory for deep learning via over-parameterization.
\newblock In {\em Proceedings of the 36th International Conference on Machine
  Learning}, volume~97, pages 242--252. PMLR, 09--15 Jun 2019.

\bibitem{Du:19}
Simon Du, Jason Lee, Haochuan Li, Liwei Wang, and Xiyu Zhai.
\newblock Gradient descent finds global minima of deep neural networks.
\newblock In {\em Proceedings of the 36th International Conference on Machine
  Learning}, volume~97, pages 1675--1685. PMLR, 09--15 Jun 2019.

\bibitem{Adavani:20}
Madhu~S. Advani, Andrew~M. Saxe, and Haim Sompolinsky.
\newblock High-dimensional dynamics of generalization error in neural networks.
\newblock {\em Neural Networks}, 132:428--446, 2020.

\bibitem{Oymak:20}
Samet Oymak and Mahdi Soltanolkotabi.
\newblock Toward moderate overparameterization: Global convergence guarantees
  for training shallow neural networks.
\newblock {\em IEEE Journal on Selected Areas in Information Theory}, PP:1--1,
  04 2020.

\bibitem{Belkin:19}
S.~Ma M.~Belkin, D.~Hsu and S.~Mandal.
\newblock Reconciling modern machine-learning practice and the classical
  bias–variance trade-off.
\newblock {\em Proceedings of the National Academy of Sciences},
  116(32):15849--15854, 2019.

\bibitem{Spigler:19}
S~Spigler, M~Geiger, S~d’Ascoli, L~Sagun, G~Biroli, and M~Wyart.
\newblock A jamming transition from under- to over-parametrization affects
  generalization in deep learning.
\newblock {\em Journal of Physics A: Mathematical and Theoretical},
  52(47):474001, 2019.

\bibitem{Neal:18}
Brady Neal, Sarthak Mittal, Aristide Baratin, Vinayak Tantia, Matthew Scicluna,
  Simon Lacoste-Julien, and Ioannis Mitliagkas.
\newblock A modern take on the bias-variance tradeoff in neural networks, 2018.

\bibitem{Neal:19}
Brady Neal.
\newblock On the bias-variance tradeoff: Textbooks need an update.
\newblock {\em arXiv}, 2019.

\bibitem{Nakkiran:21}
Preetum Nakkiran, Gal Kaplun, Yamini Bansal, Tristan Yang, Boaz Barak, and Ilya
  Sutskever.
\newblock Deep double descent: where bigger models and more data hurt.
\newblock {\em Journal of Statistical Mechanics: Theory and Experiment},
  2021(12):124003, 2021.

\bibitem{Ba:20}
Jimmy Ba, Murat Erdogdu, Taiji Suzuki, Denny Wu, and Tianzong Zhang.
\newblock Generalization of two-layer neural networks: An asymptotic viewpoint.
\newblock In {\em International Conference on Learning Representations}, 2020.

\bibitem{Krogh:91}
Anders Krogh and John Hertz.
\newblock A simple weight decay can improve generalization.
\newblock In {\em Advances in Neural Information Processing Systems}, volume~4.
  Morgan-Kaufmann, 1991.

\bibitem{Lecun:98}
Yann LeCun, Corinna Cortes, and Christopher~J.C. Burges.
\newblock The mnist database of handwritten digits, 1998.

\end{thebibliography}

\end{document}